\documentclass[sigconf]{acmart}

\usepackage{booktabs} 
\usepackage{comment}
\usepackage[titlenumbered,ruled]{algorithm2e}
\usepackage{algpseudocode}
\usepackage{graphicx}
\usepackage{array}
\usepackage{bm}
\usepackage{footnote}
\usepackage{caption}
\usepackage{subcaption}

\usepackage{amsthm,amsmath,amsfonts,amssymb,epsfig,epstopdf,url,array}
\usepackage[T1]{fontenc}
\usepackage[utf8]{inputenc}
\usepackage{colortbl}
\usepackage{widetext}
\usepackage{flushend}
\usepackage{cuted}
\usepackage{breqn}
\usepackage{multirow}
\usepackage{bbm}

\usepackage{enumitem}

\newtheorem{defn}{Definition}[section]

\DeclareMathOperator*{\argmax}{arg\,max}

\begin{document}

\copyrightyear{2018} 
\acmYear{2018} 
\setcopyright{acmcopyright}
\acmConference[CIKM '18]{The 27th ACM International Conference on Information and Knowledge Management}{October 22--26, 2018}{Torino, Italy}
\acmBooktitle{The 27th ACM International Conference on Information and Knowledge Management (CIKM '18), October 22--26, 2018, Torino, Italy}
\acmPrice{15.00}
\acmDOI{10.1145/3269206.3271681}
\acmISBN{978-1-4503-6014-2/18/10}

\fancyhead{}

\title{JIM: Joint Influence Modeling for Collective Search Behavior}

\author{Shubhra Kanti Karmaker Santu}
\affiliation{University of Illinois Urbana-Champaign (UIUC)}
\email{karmake2@illinois.edu}

\author {Liangda Li}
\affiliation{Yahoo Research}
\email{liangda@oath.com}
       
\author{Yi Chang}
\authornote {this author is also affiliated to College of Computer Science and Technology, Jilin University and Key Laboratory of Symbolic Computation and Knowledge Engineering of Ministry of Education, China}
\affiliation{College of Artificial Intelligence\\Jilin University, Changchun, China}
\email{yichang@acm.org}

\author{ChengXiang Zhai}
\affiliation{University of Illinois Urbana-Champaign (UIUC)}
\email{czhai@illinois.edu}

\begin{abstract}
Previous work has shown that popular trending events are important external factors which pose significant influence on user search behavior and also provided a way to computationally model this influence. However, their problem formulation was based on the strong assumption that each event poses its influence independently. This assumption is unrealistic as there are many correlated events in the real world which influence each other and thus, would pose a joint influence on the user search behavior rather than posing influence independently. In this paper, we study this novel problem of Modeling the Joint Influences posed by multiple correlated events on user search behavior. We propose a \emph{Joint Influence Model} based on the Multivariate Hawkes Process which captures the inter-dependency among multiple events in terms of their influence upon user search behavior. We evaluate the proposed \emph{Joint Influence Model} using two months query-log data from https://search.yahoo.com/. Experimental results show that the model can indeed capture the temporal dynamics of the joint influence over time and also achieves superior performance over different baseline methods when applied to solve various interesting prediction problems as well as real-word application scenarios, e.g., \emph{query auto-completion}.
\end{abstract}



\maketitle

\section{Introduction}\label{sec:intro}
Search Engine optimization has been a vastly studied research area in the past decade. One key component of search engine optimization is analyzing the user search behavior in order to better understand their information need. User search behavior has been studied from multiple perspectives, e.g., user's own browsing history, click log analysis etc.  Recently, how various external factors influence the user search behavior has attracted increasing attention~\cite{karmaker2017modeling}. One important type of external factor is the external events that ``significantly'' attract the general mass. They trigger user's thirst for information related to the event and thus, pose influence on how the users search to fulfill their information need. How to model the influence of such external events on user search behavior is the high level research question we study in this paper.

\begin{figure}[!htb]
	\centering
	\includegraphics[width=87.0mm]{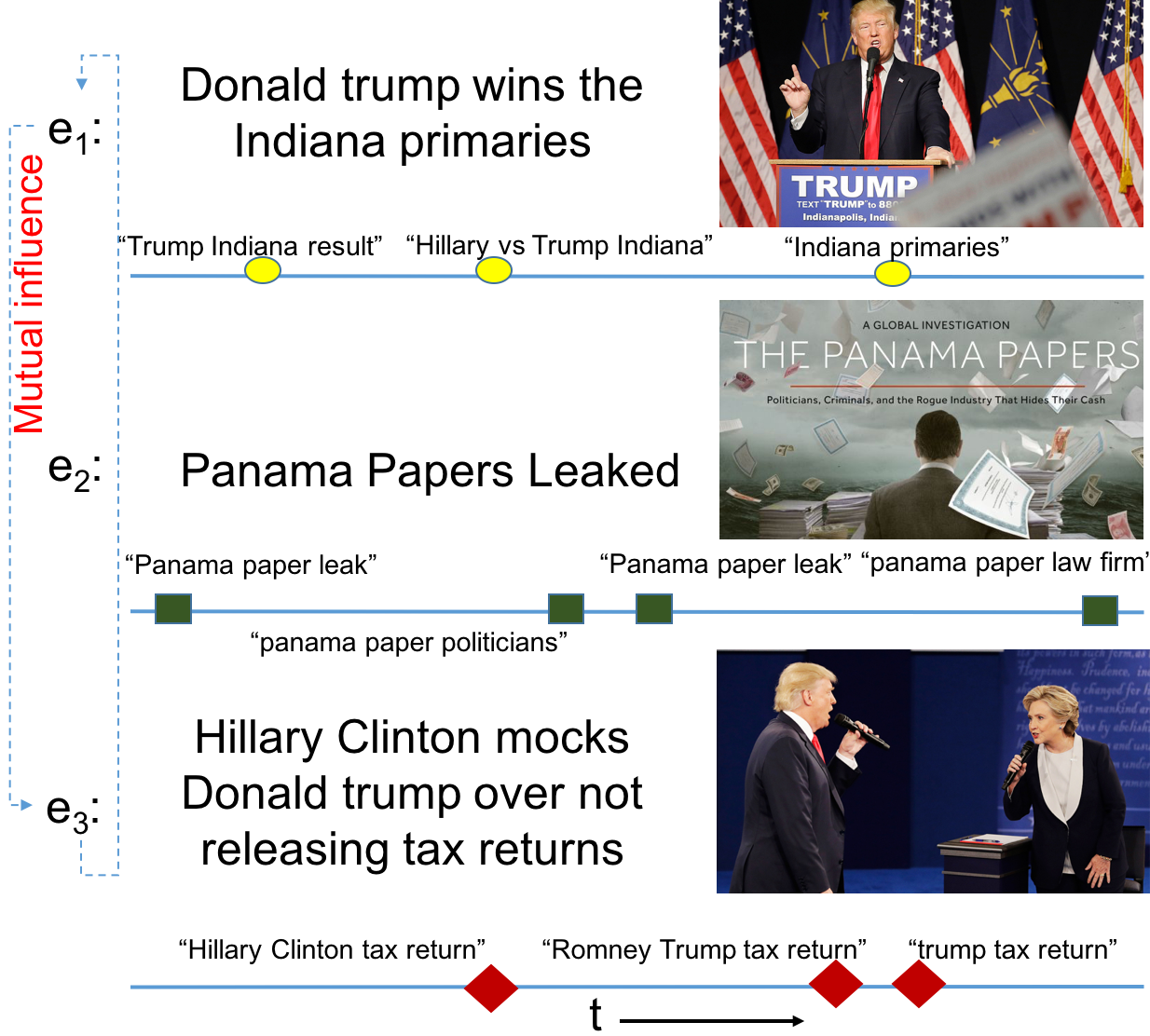}
	\vspace{-7mm}
	\caption{A toy example with three events $e_1,e_2,e_3$. The circles, squares and dices represent queries generated by the influence of event $e_1$, $e_2$ and $e_3$ respectively.}
	\label{fig:problem}
	\vspace{-7mm}

\end{figure}

The problem of modeling the influence of popular trending events on user search behavior is not entirely new, specifically, this problem was introduced by Karmaker et.al.~\cite{karmaker2017modeling}. However, the problem definition provided in~\cite{karmaker2017modeling} was based on the strong assumption that the influence posed by each event is independent of the other events, which clearly limits the applicability of such solution to cases where there are multiple correlated events and these events pose a joint influence on individual user's search pattern. To clearly motivate the problem, let us start with the example in Figure~\ref{fig:problem}, where we show three popular events from the month of April, 2016. The first event (denoted by $e_1$) is Donald Trump's win in the Indiana Primaries. The \emph{blue} line below the event description represents the time dimension and the ``yellow'' dots represent queries related to/triggered by the event $e_1$. For example, the query ``Trump Indiana result" is clearly seeking information about Trump's election results for Indiana Primaries. Note that, the same query can be posed by multiple users at different instants of time. Here, $e_1$ is an influential event that has triggered a lot of user queries related to that event. We call these triggered queries as \emph{Influenced queries}. Similarly, event $e_2$, i.e., ``Panama Papers Leaked'' and event $e_3$, i.e., ``Hillary Clinton mocks Donald Trump over not releasing tax returns'' also trigger numerous queries from users asking for relevant information about the respective event. A deeper thought would also reveal that some of these events may be correlated and they may have a joint-influence on the generation of some queries. For example, people searching for ``Hillary Clintons mocking about Donald Trump" might also be interested in information about Trump's Indiana Primary Results and vice-versa. Thus, mutual influence exist among events that jointly affect user search behavior and this joint influence also evolves over time causing corresponding change in the user search pattern. In this paper, we model this evolution of joint influence posed by multiple external events on the search behavior of users.

As mentioned in the previous paragraph, the major limitation of the previous work by Karmaker et.al.~\cite{karmaker2017modeling} is the assumption that influence posed by one event is independent of the other events. In this paper, we relax this assumption by providing a new problem formulation, i.e., modeling the joint influence posed by multiple events on user search behavior. Specifically, we introduce a new data mining problem, where, given a search query log and a set of (correlated) events, the task is to mine both these datasets to infer the joint influence posed by the provided set of (correlated) events on triggering queries from users. This specifically means, beside measuring the influence of the primary event that triggered the query (lets call it \emph{Direct Influence}), the task also requires to measure the influence of secondary (correlated) events for the same (lets call it \emph{Indirect Influence}). This is a new problem because besides computing the degree of influence posed by each event, we also need to come up with a way to compute how their influences are temporally correlated to each other. The joint influence mining task naturally raises many associated interesting research questions, including, how to come up with a numerical formula for measuring influence that is comparable across multiple events (note that, the influence scores computed by Karmaker et.al.~\cite{karmaker2017modeling}  are not directly comparable across multiple events), how influence of multiple events jointly evolve over time and how they correlate in the temporal dimension etc. (see section~\ref{sec:problem} and section~\ref{sec:questionsquality} for a detailed list of questions).

To solve the joint influence modeling task, we propose a novel mining algorithm based on Multivariate Hawkes Process~\cite{liniger2009multivariate}, which is a mutually exciting point process suitable for modeling the frequencies of random events. The joint influence modeling approach proposed by us has several benefits over the independent influence model proposed in~\cite{karmaker2017modeling}; first, it relaxes the assumption that each event poses an influence that is independent of the other events and thus can model real word scenarios better; second, it can capture the temporal correlation of influences posed by two correlated events providing a way to categorize direct influence versus indirect influence and thus can leverage this correlation to better model the evolution of joint influence over time; third, it provides a formal way to measure the influence of multiple events in a comparable numerical scale. Another beneficial feature of the proposed method, as demonstrated by the experimental results (section~\ref{sec:results}), is that the proposed joint influence model is fairly general and is widely applicable on various interesting prediction tasks and search intent related applications (e.g., query suggestion, query auto-completion) and obtains superior results in comparison to multiple baseline methods. The core contributions of this paper are listed below:

\vspace{-1mm}
\begin{enumerate}[leftmargin=.2cm,itemindent=.5cm,labelwidth=\itemindent,labelsep=0cm,align=left]
\item We introduce the novel problem of modeling the (temporal) dependency across multiple events in terms of the influences posed by them on user search behavior. To the best of our knowledge, this problem has not been studied before.

\item We propose a \emph{Joint Influence Model} based on \emph{Multivariate Hawkes Process} which captures the joint-influence posed by multiple events on user search behavior as well as models how this joint influence evolves over time.

\item We present efficient numerical techniques to compute the likelihood of any query log data w.r.t. the  proposed \emph{Joint Influence Model}; which provides us with a way to estimate the optimal parameters of the model by maximizing the likelihood.

\item We evaluate the proposed \emph{Joint Influence Model} using two months query-log data from https://search.yahoo.com/. Experimental results show that the model can indeed capture the temporal dynamics of the joint influences over time and can be applied to solve various interesting prediction problems as well as real-word application scenarios, e.g., \emph{query auto-completion}.
\end{enumerate}

\section{Related Work} \label{sec:related_work}

Search query logs have been extensively studied to understand user search behavior and provide better search experience \cite{jiang2014searching,li2014identifying,white2013enhancing}. Existing work mostly focused on the inference of users' search intent based on their own search habit and search history. On the other hand, our paper tries to model how user behavior on a search engine is influenced by external factors such as trending events.

Temporal Information Retrieval~\cite{campos2015survey,dakka2012answering,berberich2010language,kulkarni2011understanding} and Event Detection~\cite{atefeh2015survey,zhou2015unsupervised,walther2013geo,dong2015multiscale,abdelhaq2013eventweet} are two areas closely related to our work. While Event Detection has been studied vastly in the literature (see~\cite{atefeh2015survey} for a recent survey), research interest on Temporal Information Retrieval has grown recently~\cite{campos2015survey}. However, we emphasize that, neither of these is the intended goal of this study and our primary motivation is somewhat orthogonal, i.e., given that some (possibly multiple) events have already been reported, we go one step further to investigate how these events may jointly impact/influence the search behavior of the users.

The notion of event-based retrieval was introduced by Str\"{o}tgen and Gertz~\cite{strotgen2012event} by returning events instead of documents. Zhang et al.~\cite{zhang2010learning} addressed the detection of recurrent event queries. Ghoreishi and Sun~\cite{ghoreishi2013predicting} introduced a binary classifier for detecting queries related to popular events. Kanhabua~\cite{kanhabua2015learning} extended the work~\cite{ghoreishi2013predicting} by enabling the classifier to detect less popular queries beside popular ones. However, all these approaches are supervised classification methods and largely depend on the quality of training labels provided by humans, whereas our approach is unsupervised.

Kairam et. al.~\cite{Kairam:2013} investigated the online information dynamics surrounding trending events, by performing joint analysis of large-scale search and social media activity. Matsubara et. al.~\cite{matsubara2015web}  presented a new model for mining large scale co-evolving online activities. Pekhimenko et al. \cite{Pekhimenko:2015} designed a system named ``PocketTrend" that automatically detects trending topics in real time, identified the search content associated to the topics, and then intelligently pushed this content to users' local machine in a timely manner. However, none of these studies provide answer to the question: how to model the evolution of joint influence posed by multiple events on user search behavior, which is one of the primary motivations of our work. The closest match to this paper is the work by Karmaker et.al. \cite{karmaker2017modeling} where they first introduce the problem of modeling the influence of popular trending events on user search behavior. However, as mentioned in section~\ref{sec:intro}, their problem definition was based on the unrealistic assumption that only one event can influence the triggering of a particular query and the influences posed by multiple events are independent of each other. In this paper, our primary focus is to relax these assumptions and propose a more realistic model to capture the joint influence of multiple events.

Another important topic related to this paper is point process, which has been used to model social networks~\cite{Blundell:2012} and natural events~\cite{Zhuang:2002}. People find self-exciting point processes naturally suitable to model continuous-time events where the occurrence of one event can affect the likelihood of subsequent events in the future. One important self-exciting process is Hawkes process, which was first used to analyze earthquakes \cite{Zhuang:2002}, and then widely applied to many different areas, such as market modeling \cite{Errais:2010}, crime modeling \cite{Stomakhin:2011}, conflict \cite{Zammit-Mangion:2012}, viral videos on the Web \cite{Crane:2008} etc. In this work, we propose a novel Joint Influence Model based on multivariate Hawkes process~\cite{liniger2009multivariate} that can capture the dynamics of simultaneous influence by multiple events on user search behavior.

\section{Problem Formulation}\label{sec:problem}

Let, $E=\{e_1,e_2, ..., e_k \}$ be the set of all events for which we want to analyze their influence on the user search behavior, where $k$ is the total number of events under consideration and each event $e_j$ is represented in terms of natural text (for details on the representation of an event, refer to the work by Karmaker et.al.~\cite{karmaker2017modeling}). Also assume that, each $e_j$ is associated with a set of queries that were generated from influence (``to some extent'') by the same event. Let this set be denoted by  $Q_j=\{q_{j1},q_{j2}, ....\}$. Each $q_{ji}$ consists of a tuple <$w_{ji},t_{ji},x_{ji}$>, where, $w_{ji}$ is the query-text, $t_{ji}$ is the timestamp of receiving the query and $x_{ji}$ is a textual-similarity score between event-text $e_j$ and query-text $w_{ji}$. The higher the similarity between $e_j$ and $w_{ji}$, the higher the $x_{ji}$ score is. For details on how we can get the query set $Q_j$ associated with each event $e_j$ and how to compute $x_{ji}$ for an event-query pair, please see~\cite{karmaker2017modeling}. We omit the details here due to lack of space.

Given the input data mentioned above, our goal is to model the temporal dynamics of the joint influence posed by different events in $E$ on user search behavior. Specifically, we seek answers to the following questions which were never investigated before: 1) Is there a way to computationally model the dependency among different events in terms of the influences posed by them on user search behavior? 2) How these (correlated) influences of multiple events jointly evolve over time? 3) Given that we have seen a query which is triggered by some event $e_j$, how does that change the future influence of some event \emph{other} than $e_j$? 4) Can we use the correlation among multiple events to distinguish between \emph{Direct Influence} and \emph{Indirect Influence} (defined in Section~\ref{sec:intro})? We also ask the same questions raised by Karmaker et.al.~\cite{karmaker2017modeling}, e.g., 5) How the textual similarity between an influential event and an influenced query affects the influence trend of that event?  6) How long the influence of different events last?  
To provide answers to these questions, we formally introduce a novel \emph{Joint Influence Model} based on Multivariate Hawkes Process, in the following section. 


\section{Joint Influence Model}

We model the joint influence of multiple events on user search behavior through a generative multivariate point process where each point corresponds to the submission of a new query influenced by some event $e_j \in E$. To be more specific, we propose a new generative model based on Multivariate Hawkes Process (a specific mutually exciting point process) to describe the generation of the influenced queries. This way of modeling query generation is beneficial because this would also allow us to quantify the influence of different events on this generation process at any instant of time. Multivariate Hawkes process is naturally suitable to our problem scenario because it can model the frequencies of occurrences of multiple events in the continuous time domain. For a detailed background on Multivariate Hawkes Process and for further justification on why it is helpful, please refer to~\cite{liniger2009multivariate}.

Let, $Q=Q_1\cup Q_2 \cup .... \cup Q_k$, be the set of all query submissions which were influenced by some event $e_j \in E$. Additionally, let $Q_j$ be the set of all queries that were triggered by the direct influence of event $e_j$. One naive way to collect $Q_j$ corresponding to event $e_j$ is to retrieve queries from the search log that are textually similar to event-text. For further details on how to retrieve a good quality $Q_j$ for event $e_j$, please refer to~\cite{karmaker2017modeling}. For modeling the joint influence, we consider the union set, i.e., $Q$, where each query $q_i \in Q$ corresponds to one point in the multivariate point process and is represented by the tuple <$t_i$,$d_i$,$x_i$>. Here, $t_i$ is the timestamp of receiving the the query and thus, always $t_i>0$; $d_i$ is the event which influenced the generation of $q_i$ and thus, $d_i$ can be any event $e_j$, i.e., $d_i \in E$;  $x_i$ is the textual-similarity score between event-text, $text(d_i)$ and query-text, $text(q_i)$. 

Given this setup, the core technical challenge in designing the \emph{Joint Influence Model} boils down to the problem of how we can formally define the multi-event influenced query generation process; in other words, how to fully characterize the multivariate point process? This is not trivial due to the abstractness in the concept of influence. We address this challenge by introducing the notion of \emph{Influence Function}, which we will discuss in detail in the following section\footnote{All the codes and evaluation scripts for experimentation can be found at the following link: ({\bf https://bitbucket.org/karmake2/influencemodeling/src/master/})}:

\vspace{-1mm}
\subsection{The Influence Function}\label{sec:influenceFunction}
We characterize the multivariate point process by defining a set of continuous functions $\lambda_j$ for $j=\{1,2,....,k\}$, we call them \emph{Influence Functions}, which represent the influence of each event $e_j \in E$ on the generation of the queries in $Q$ at any instant of time. Designing a suitable $\lambda$ function is the main challenge towards building a reasonable Joint Influence Model. However, defining influence is more of a philosophical question rather than a mathematical one. With this constraint in mind, we adopt to define influence through different components that the final influence function should accommodate and eventually, combine all these components into a single influence function. 
We first start with various components of the influence function $\lambda_j$.\\

\noindent {\bf Base Influence:} We assume that there is always a non-negative influence posed by each event $e_j \in E$ on the generation of the queries in $Q_j$. Thus, each event $e_j$ is associated with a constant $\eta_j$ which governs the rate at which we expect to observe new queries influenced by event $e_j$. This gives our first set of parameters for the influence function, i.e., $\eta_j \ge 0$  for $j=\{1,2,....,k\}$. In contrast, the independent influence model proposed by Karmaker et.al.~\cite{karmaker2017modeling} (let's call it \emph{IIM}), has only one parameter $\eta$ for all events.\\

\noindent {\bf Decay Functions:} The decay functions characterize how the influence of each event diminishes over time. Thus, each event $e_j$ is associated with a decay function $w_j$ which decides how fast the influence of the same event decays with time. Without loss of generality, we use Exponential Decay functions for our \emph{Joint Influence Model}. While other forms of the decay function are certainly possible, the investigation of the choosing the right decay function is orthogonal to the goal of this research. 
Mathematically, Exponential Decay Functions are represented as the following:

\vspace{-2mm}
\[ w_j(t)=\alpha_j \exp(-\alpha_jt)\]
\vspace{-4mm}

The corresponding cumulative decay function is the following (we will need this later):

\vspace{-4mm}
\[ \bar w_j(t)=1- \exp(-\alpha_jt)\]
\vspace{-4mm}

In contrast, \emph{IIM}~\cite{karmaker2017modeling} has one decay parameter $w$ for all events.\\

\noindent {\bf Impact Functions:} Whenever the search engine receives a query triggered by some event $e_j$, we assume that this newly received query increases the influence of all events (not only $e_j$), which in turn, increases the probability of receiving further queries influenced by different events. The more we receive new (influenced) queries, the higher the influence of different events become; yielding a higher probability of receiving more influenced queries in the future. Thus, the influence of different events as well as frequency of influenced queries we receive mutually grow together, which is similar to the idea of mutually exciting multivariate point processes~\cite{hawkes1971spectra}. Note that, for some event (mostly uncorrelated events), the increment of its influence can be zero which is also expected.

Given that we have received a new query $q_i$ (<$t_i$,$d_i$,$x_i$>) triggered by event $d_i$, the amount by which the influences of different events increase depends on the textual-similarity score, i.e. $x_i$, between the event-text and the query-text. This is intuitive because, highly relevant queries are expected to have more impact on the change of influence than less relevant queries. To capture this, we introduce a set of \emph{Impact Functions} which govern how the influence of all events change depending on the textual-similarity score between the newly received query and its triggering event.  Let us denote these \emph{Impact Functions} by the notation $g_{d_i}(x_i)$. The interpretation of $g_{d_i}(x_i)$ is as follows: assuming that the newly received query $q_i$ was triggered by event $d_i$ and the textual similarity between $text(d_i)$ and $text(q_i)$ is $x_i$, the influences of all the events are then increased in proportion to $g_{d_i}(x_i)$.

Note that, reception of query $q_i$ increases the influences of all events by the same amount, i.e., by $g_{d_i}(x_i)$: which is not desirable. To address this issue, we have a whole new set of parameters, namely ``Mutual-Influence Co-efficients" which we will discuss shortly after this. However, the purpose of ``Impact Functions'' is solely to define how the influence of an event changes based on the textual-similarity score between the newly received query $q_i$ and its triggering event $d_i$.

Impact functions take the textual similarity score $x_i$ as an input parameter. The exact form of Impact Function we choose would thus depend on the distribution of $x_i$, let us call it \emph{Intent-Match Distribution}. Below we discuss the \emph{Intent-Match Distribution} briefly, choose a reasonable function for it and then choose the corresponding suitable \emph{Impact Function}\\.

\noindent {\bf Intent-Match Distribution:} Intent-Match Distribution is essentially the distribution of the textual-similarity score between the triggering event and the influenced query. For textual-similarity score, we choose the following modified version of the $BM25$ introduced in the work~\cite{karmaker2017modeling} (The details of this function and rationale behind choosing it can be found in the paper~\cite{karmaker2017modeling} ). Let, $W_E=<W_{E_1}, W_{E_2}, ... , W_{E_n}>$ be the ``event-text'' and $W_q=<W_{q_1}, W_{q_2}, ...,  W_{q_n} >$ be the ``query-text''. Then,

{
\[ x_i(W_E,W_q)=\sum_{i=1}^{|W_E|}   \frac{\omega(W_{E_i}).IDF(W_{E_i}). TF(W_{E_i}, W_q).(k_1+1)} {TF(W_{E_i}, W_q)+ k_1.(1-b+b.\frac{|W_q|}{avgql})} \]
\vspace{-3mm}
\begin{equation} \label{equ:similarity}
\text{subject to}  \sum_{i=1}^{|W_E|}  \omega(W_{E_i})=1
\end{equation}
\vspace{-2mm}
}

Note that, the textual-similarity score $x_i$ is independent of the past history of received queries and solely depends on the similarity between the ``event-text" and the ``query-text". Further, $x_i\ge 0$.

To specify the \emph{Intent-Match Distribution}, we hypothesize that a power law probability distribution is the most suitable for our case because of the following reasoning: among the set of all queries that are influenced by some event $e_j$, very few queries would exactly match with the details in the event-text, while a lot of queries intent would match the details only partially or marginally (these are general exploratory queries). The higher the intent-match, the rarer the frequency becomes; in fact, the frequency decreases exponentially with the increase in textual-similarity. Our empirical evaluation also supports this hypothesis (details in section~\ref{sec:resultsquality}).  

Based on the argument presented above and without loss of generality, we select ``Pareto distribution" as our \emph{Intent-Match Distribution}, which is a popular power law probability distribution. 
``Pareto distribution" is defined on the half line $[0,\infty)$ and has two parameters $\mu>0$ and $\rho>0$. Each event $e_j$ is associated with a Intent-Match Distribution $f_j$ (``Pareto distribution'' in this case).

\vspace{-2mm}
\begin{equation}\label{equ:intentDistribution}
f_j(x)=\frac{\rho_j\mu_j^{\rho_j}}{(x+\mu_j)^{\rho_j+1}}
\end{equation}

Under the restriction that $\rho_j > 2$, a suitable impact function is the following with parameters $\rho_j \ge 0$, $\mu_j \ge 0$, $\phi_j \ge 0$, $\psi_j \ge 0$ (Please see~\cite{liniger2009multivariate} for details and rationale) :

\begin{equation}\label{equ:impactFunction}
g_j(x)=\frac{(\rho_j-1)(\rho_j-2)}{\phi_j(\rho_j-1)(\rho_j-2)+\psi_j\mu_j(\rho_j-2)}(\phi_j+\psi_jx)
\end{equation}

Thus, each event $e_j$ is associated with a Intent-Match Distribution $f_j$ as well as an impact function $g_j$. In contrast, the \emph{IIM}~\cite{karmaker2017modeling} has only one impact function $g(x)$ for all events, which was defined as $g(x)=x$; whereas, $g_{d_i}(x_i)$ is a generalization of that with more flexibility to capture the impact.\\


\vspace{-1mm}
\noindent {\bf Mutual-Influence Co-efficients:} While the impact function captures the relationship between the textual-similarity of an ``influenced query-triggering event pair'' and the corresponding change in the influence of an event, it fails to distinguish the different impacts the same query might pose for different events. For example, the submission of query ``Trump Indiana Results'' should directly indicate an increasing influence of the event ``Donald Trump wins Indiana primaries'' (This is the \emph{Direct Influence}); however, the same query might have little/no indication about the increasing influence of the event ``Messi scores a hat-trick against real madrid'' (lets call it \emph{No Influence}). At the same time, query ``Trump Indiana Results'' might have an indirect indication about the increasing influence of the correlated event ``Hillary Clinton results for Iowa Primaries'' (The is the \emph{Indirect Influence}). Modeling these inter-dependencies among multiple events in terms of the influence posed by them is one of the central key questions we investigate in this paper. Our proposed \emph{Joint Influence Model} addresses this question by introducing a new set of Co-efficients, we call them Mutual-Influence Co-efficients, which is a unique component of our proposed model.

To capture the three types of influences, i.e., \emph{Direct Influence, Indirect Influence, No Influence} as mentioned above; we introduce a $k\times k$ matrix of coefficients which we call the Mutual-Influence Co-efficients. 

\vspace{-5mm}
{

\[
MIC=
\begin{bmatrix}
    \nu_{11}       & \nu_{12} & \nu_{13} & \dots & \nu_{1k} \\
    \nu_{21}       & \nu_{22} & \nu_{23} & \dots & \nu_{2k} \\
    \hdotsfor{5} \\
    \nu_{k1}       & \nu_{k2} & \nu_{k3} & \dots & \nu_{kk}
\end{bmatrix}
\]
}

The diagonal elements of the matrix represent \emph{Direct Influence}, while non-diagonal elements represent \emph{Indirect Influence}. We also impose the constraint, $\nu_{ji}\ge 0$ for $i,j=\{1,....,k\}$. A zero value for any element in the $MIC$ matrix represents \emph{No Influence}, while higher non-zero values indicate \emph{Significant Influence}. Thus, the $MIC$ matrix contains valuable information about the inter-dependencies among multiple external events in terms of their influence on user search behavior.

\noindent {\bf Influence Function and Query Generation Process:} 
So far, we have discussed all the components we needed to define our influence function. Below we present the 
actual definition of the influence function by combining all these components: 

\vspace{-4mm}
\begin{equation}\label{equ:influence}
\lambda_j(t)=\eta_j+\sum_{j=1}^k \nu_{ji}  \int_{(-\infty,t)\times R} w_j(t-s)g_j(x) e_j (ds\times dx)
\end{equation} 

Now, we define the mutually-exciting query generation process:

\begin{defn}[Mutually-Exciting Query Generation Process] \label{def:pointProcess}
Let us assume that, we observe queries in the form of triples <$t_i$,$d_i$,$x_i$> for $1\le i \le n$, where $t_i \in [T_*,T^*]$ and $t_i>t_{i-1}$, $d_i\in \{1,2,....,k\}$ and $x_i \in R^{+}$. For the $i$-th query, it occurs at timestamp $t_i$, the triggering event is $d_i$ and the corresponding textual-similarity is $x_i$. At any instant of time $t$, each event $e_j$ for $j=\{1,2,.....,k\}$ has an influence 
$\lambda_j$ defined by equation~\ref{equ:influence}. This constitutes our Generative Multivariate Hawkes Process.
\end{defn}

For a Multivariate Hawkes Process to be well defined, we need the following two conditions to be satisfied:
\begin{enumerate}
\item The maximum of the Eigen Values of the $MIC$ matrix is defined as the spectral radius of $MIC$, i.e, \\$Spr(MIC)=max(eigenValues(MIC))$. Multivariate Hawkes Process requires the following condition to satisfy:
\vspace{-1mm}
\[Spr(MIC)<1\]
\vspace{-5mm}
\item The decay functions must satisfy the following constraints:
\vspace{-1mm}
\[\int_0^{\infty} tw_j(t) dt < \infty\]
\end{enumerate}

Finally, for computational feasibility, we present the numerical version of the continuous influence function in equation~\ref{equ:influence} below. Let us assume that we have observed queries at points $\{t_i\}$, for $1\le i\le n$. Then, for any timestamp $t_i$, the influence of event $j$, $\lambda_j(t_i)$ is defined as: 

\vspace{-5mm}
\begin{equation}
{\hat{\lambda}_j(t_i)}=\eta_j+\sum_{m=1}^{i-1} \nu_{j,d_m} w(t_i-t_m) g_{d_m}(x_m)
\end{equation}

\subsection{Estimation of the Optimal Parameters}\label{sec:estimation}
This section presents the estimation techniques for the optimal parameter values of the influence function. For this purpose, we define the likelihood function for any observed sequence of queries with respect to the proposed mutually exciting multivariate point process. We find the optimal parameters by maximizing the likelihood of the observed query data. Specifically, the log-likelihood function corresponding to the \emph{Mutually-Exciting Query Generation Process} (see Definition~\ref{def:pointProcess}) looks like the following:

\vspace{-2mm}
{
\begin{dmath}\label{equ:loglikelihood}
\log L=\sum_{j=1}^d \int_{[T_*,T^*]\times R} \log \lambda_j(t)e_j(dt\times dx) + \sum_{j=1}^k \int_{[T_*,T^*]\times R} \log f_j(x) e_j(dt\times dx) -\sum_{j=1}^k \Lambda_j(T^*)
\end{dmath}
}

Here, $T^*$ is the upper bound of the observation period and $\hat{\Lambda}_j(T^*)$ is called  the compensator function and is defined as follows:

{
\vspace{-1mm}
\begin{dmath}\label{equ:compensator}
\Lambda_j(t)=\eta_j(t-T_*)+\sum_{m=1}^k \nu_{jm} \int_{(-\infty,t)\times R} [\hat{w}_j(t-u)-\hat{w}_j(T_*-u)]g_m(x) e_m(du\times dx)
\end{dmath}
}

\noindent{\bf Numerical Computation:} For computational feasibility, we now present the way to numerically compute the log-likelihood function defined in Eqn~(\ref{equ:loglikelihood}). Specifically, the numerical version of the log-likelihood function takes the following form:

{
\vspace{-5mm}
\begin{equation}\label{equ:loglikelihoodnumeric}
\log \hat{L}=\sum_{i=1}^n \log \hat{\lambda}_{d_i}(t_i) + \sum_{i=1}^n \log f_{d_i} (x_i) - \sum_{j=1}^k \hat{\Lambda}_j(T^*)
\end{equation}
}

While computation of $f_{d_i} (x_i)$ is straight-forward from equation~\ref{equ:intentDistribution}, computation of $\hat{\lambda}_{d_i}(t_i)$ and $\hat{\Lambda}_j(T^*)$ are more involved. Below we present the exact formulas to compute $\hat{\lambda}_{j}(t_i)$ and $\hat{\Lambda}_j(T^*)$ omitting the derivation details due to lack of space. We assume exponential decay function, i.e., $w_j=\alpha_j\exp(-\alpha_jt)$, for the exact computational formula, while other forms of decay functions are certainly possible.

{\vspace{-2mm}
\begin{dmath}\label{equ:numericIntensity}
\hat{\lambda}_{j}(t_i)=\eta_j + [\lambda_j(t_{i-1})-\eta_j]\exp[-\alpha_j(t_i-t_{i-1})]+\nu_{j,d_{i-1}}g_{d_{i-1}}(x_{i-1})\alpha_j\exp[-\alpha_j(t_i-t_{i-1})]
\end{dmath}

\begin{dmath}\label{equ:numericCompensator}
\hat{\Lambda}_j(T^*)=\eta_j(T^*-T_*)+\sum_{i=1}^n \nu_{j,d_i} \bar{w}_j(t^*-t_i) g_{d_i} (x_i)
\end{dmath}
}

By plugging in equation~\ref{equ:intentDistribution},~\ref{equ:numericIntensity} and~\ref{equ:numericCompensator}, we obtain the complete numerical version of the log-likelihood function as follows:

{\vspace{-2mm}
\begin{dmath}\label{equ:loglikelihoodnumericFinal}
\log \hat{L}=\sum_{i=1}^n \log \left\{\eta_j + [\lambda_j(t_{i-1}-\eta_j)]\exp[-\alpha_j(t_i-t_{i-1})]+\nu_{j,d_{i-1}}g_{d_{i-1}}(x_{i-1})\alpha_j\exp[-\alpha_j(t_i-t_{i-1})]\right\} + \sum_{i=1}^n \log  \left(\frac{\rho_{d_i}\mu_{d_i}^{\rho_{d_i}}}{(x+\mu_{d_i})^{\rho_{d_i}+1}}\right) - \sum_{j=1}^k \left\{\eta_j(T^*-T_*)+\sum_{i=1}^n \nu_{j,d_i} \bar{w}_j(t^*-t_i) g_{d_i} (x_i)\right\}
\end{dmath}
}

Here, $g_{d_i} (x_i)$ is defined by as:
\[g_{d_i}(x)=\frac{(\rho_{d_i}-1)(\rho_{d_i}-2)}{\phi_j(\rho_{d_i}-1)(\rho_{d_i}-2)+\psi_{d_i}\mu_{d_i}(\rho_{d_i}-2)}(\phi_{d_i}+\psi_{d_i}x)\]

Given the log-likelihood function in equation~\ref{equ:loglikelihoodnumericFinal}, the set of parameters associated with it is the following:

\vspace{-3mm}
\begin{equation}
\Theta=\left\{\eta_j , \alpha_j, \nu_{ji},\rho_j,\mu_j, \phi_j, \psi_j\right\} \text{, where $(1\le i,j\le k)$}
\end{equation}

Incorporating L2 regularization, the optimization problem to find the optimal parameter set $\Theta^*$ is written as follows:

\vspace{-2mm}
\begin{equation}\label{equ:loglikelihood4}
\Theta^*=\argmax_{\Theta} \left(\log \hat{L}(\Theta)-||\Theta||\right)
\end{equation}

Here, $||\Theta||$ is the L-2 norm of the parameter vector $\Theta$. One can use any non-linear optimization method to solve this maximization problem. Nelder-Mead Simplex Method~\cite{glaudell1965nelder} is one such popular optimization technique. Another useful approach is the Sequential Least SQuares Programming (SLSQP)~\cite{boggs1995sequential}.

\section{Experimental Design}

\subsection{Data-set}
\vspace{-1mm}
Due to the absence of any readily available joint event-query dataset, we decided to create one from two sets of available data-sets: one for popular events and one for user query history. We call these two data sets \emph{Event} dataset and \emph{Query-Log} dataset respectively. The following two paragraphs provide details about these two data-sets.

{\bf Event data-set}: An obvious choice for a text data set describing events is news articles (though other data such as social media might also be applicable). The NYTimes Developers Network (thanks to them) provides a very useful api called \emph{``The Most Popular API"} \cite{NYTimes}, which automatically provides the \emph{url's} of the most e-mailed, most shared and most viewed articles from NYTimes.com during the last month from the date of the issue of the query. We chose to use this API because of two major benefits: 1) it automatically removes duplicate articles, thus we don't need to deal with cases where multiple articles are related to the same event. 2) it only provides the most popular articles from NYTimes, thus the quality/accuracy of the events represented by these articles is very high. Using this API, we collected the most e-mailed, most shared and most viewed articles for the month: April, 2016. Each article consists of a tuple <\emph{title-text}, \emph{body-text}, \emph{timestamp}>. Among different categories of news, we used four categories for our experiments: \emph{US (National Affairs)}, \emph{Movies}, \emph{Sports} and \emph{World (International Affairs)}. 

{\bf Query-Log data-set}: To analyze the user queries contemporary to the articles in \emph{Event} data-set, we use the two-months (April and May, 2016) user query log data from the widely used search engine at https://search.yahoo.com/. Each query submission $q$ is represented as a tuple <$query$-$text$, $timestamp$>. The two-months query log data contains $105,925,732$ query submissions in total.

{\bf Query-Event Joint data-set}: To create the Query-Event Joint data-set, for each article $e_j$ in the \emph{Event} data-set, we retrieved top relevant queries that have at least a similarity score of $1.25$ (with respect to $e_j$) according the textual similarity function in equation~\ref{equ:similarity} and discarded the rest. This filtering step is reasonable because if the textual similarity is very low (less than $1.25$), we assume that there is no influence of $e_j$ on the query. 
This process provides us with a set of influenced queries triggered by each event from the \emph{Event} data-set. The summary of this data-set is presented in Table~\ref{table:JointDataset}.

\begin{table}
	\begin{center}
		\begin{tabular}{|c|c|c|c|c|c|}\hline
			{\bf Section} & \bf{Total } & {\bf Avg. } & {\bf Avg. } &  \bf{Total } & {\bf Avg. }\\
			& \bf{\# of } & {\bf Title } & {\bf  Body } &  \bf{\# of } & {\bf Textual}\\
			{\bf } & \bf{events} & {\bf Length} & {\bf Length} & \bf{queries} & {\bf Sim.}\\\hline
			Movies  & 25 & 18.88 & 458.08 & 193,282 & 2.49\\
			Sports  & 15 & 19.53 & 508.4 & 616,449 & 2.48\\
			US & 18 & 20.38 & 487.77 & 204,926 & 1.99\\
			World  & 11 & 18.18 & 438.81 & 22,197 & 1.96\\\hline
		\end{tabular}
	\end{center}
	\vspace{-1mm}
	\caption{Description of Event-Query Joint Dataset}
	\label{table:JointDataset}
	\vspace{-10mm}
\end{table}

\subsection{Qualitative Evaluation of the Model}\label{sec:questionsquality}\vspace{-1mm}
It is not possible to do a direct quantitative evaluation of the influence model due to the lack of ground truth information. Thus, to evaluate the quality of the proposed \emph{Joint Influence Model}, we do a formal investigation of the optimal parameters learnt through the optimization process as described in section~\ref{sec:estimation}. Below, we present the specific research questions we ask to evaluate the model quality and provide the roadmap of how we can answer each question.

\noindent{\bf Research Questions:} 
\begin{enumerate}[leftmargin=.2cm,itemindent=.5cm,labelwidth=\itemindent,labelsep=0cm,align=left]
\item{\bf Is the ``Query Generation Process'' well-defined ?}\\
The ``Query Generation Process'' is well defined only if the Spectral Radius of the Mutual-Influence Coefficient Matrix is less than 1, i.e, $Spr(MIC) < 1$. [see section \ref{sec:influenceFunction} for more details] 

\item {\bf How to compare influences posed by different events?}\\
We can answer this question by computing average influence posed by each event and then compare them. 
The average influence vector where each element is the average influence of the corresponding event can be obtained using the following formula: $(\mathbbm{1}_k-MIC)^{-1}\boldsymbol{\eta}$, where,  $\mathbbm{1}_k$ is a $k\times k$ identity matrix.

\item {\bf How to compare \emph{Direct} Vs  \emph{Indirect} influence ?}\\
The diagonal elements of matrix $MIC$ represent the \emph{Direct Influence}, whereas, the non-diagonal elements present \emph{Indirect Influence}. We can do direct numeric comparison here.

\item {\bf How to measure the influence longevity of an event?}\\
The $\boldsymbol\alpha$ parameter defines how fast the influence of any event decays over time. Higher values of $\alpha$ denotes a faster decay.

\item {\bf Is ``Pareto Dist." suitable for ``Intent Match Dist."?}\\
To answer this question, we look at the empirical distribution of ``Intent Match" score between event-text and query text and verify whether ``Pareto Distribution" is a good match for it.

\item {\bf How well the Model fit the original data?}\\
This question can be answered by jointly plotting the simulated influence of an event and the actual frequency of queries generated by that event over the same period of time and see if the trend of the simulated influence is similar to the trend of the actual frequency of generated queries.
\end{enumerate}

\subsection{Applications and Quantitative Evaluation}\label{sec:applicaitons}\vspace{-1mm}
In this section, we demonstrate the wide applicability of the proposed ``Joint Influence Model'' by demonstrating how the model could be used to solve various interesting prediction problems as well as real-world problems associated with search engine systems. Another benefit of these experiments is to conduct indirect quantitative evaluation of the \emph{Joint Influence Model} as direct evaluation is impossible due to the lack of ground truth data for influence which is an abstract concept. The primary purpose of these experiments is to see if modeling the influence inter-dependencies among multiple events actually help us achieve better performance in real life application scenarios. To achieve these goals, we present a set of prediction tasks / application scenarios and provide a roadmap on how we can adopt the ``Joint Influence Model' to solve these tasks.\\

\noindent{\bf Application Tasks:}\vspace{-1mm}
\begin{enumerate}[leftmargin=.2cm,itemindent=.5cm,labelwidth=\itemindent,labelsep=0cm,align=left]

\item {\bf Predict the most influential event in the future: }\\
We assume the influence of an event in the current hour is proportional to the frequency of queries generated by it in the next hour. Thus, the event with the highest influence score in the current hour is predicted to be the event that generates highest number queries in the next hour. We then compare this predicted most influential event with the actual most influential event (computed from the original query log) and based on that, we can report the accuracy of the prediction for a separate held-out testing set.\\

\item {\bf Rank multiple events based on their future influences: }\\
This prediction problem is similar to previous prediction problem, except that, now we want to predict the ranking of events in terms of their future influence instead of just predicting the future top influential event. Again, we use the current hour influence scores to predict the next hour's generated query frequencies and rank the events accordingly. To evaluate the quality of the ranking, we compare the predicted ranking against the actual ranking obtained from the query log and compute two different popular ranking evaluation metrics: i.e, NDCG~\cite{wang2013theoretical} and Rank Biased Overlap (RBO)~\cite{webber2010similarity}.\\

\item {\bf Predict the most frequent query in the future: }\\
This prediction problem is the same as the prediction problem in (1) except that now we want to predict the most frequent query in the future instead of the most influential event. For this prediction task, we use a slightly modified version of the original ``Joint Influence Model'' where apart from computing the evolving influence at the event level, we also compute the evolving influence at the query level. The basic idea is to break each event-level influence into smaller units where each unit would correspond to the query level. We omit the full details of process due to lack of space.\\

\item {\bf Rank queries based on their future frequencies:}\\
This prediction problem is similar to the prediction problem in (2) except that now we want to rank queries instead of events. Again we report NDCG~\cite{wang2013theoretical} and Rank Biased Overlap(RBO)~\cite{webber2010similarity} to evaluate the quality of the predicted ranking.\\

\item {\bf Solve a real world application problem, e.g., query auto completion task:}\\
Finally, we select \emph{Query Auto Completion} as a goal task and use our proposed ``Joint Influence Model'' to solve it. Specifically, for a new query from the testing set, we look at the first word and try to predict the exact query based on the latest available influence scores of all the queries starting with the first word. Based on these influence scores, we rank the potential queries and then, compute the reciprocal rank of the actual query in the predicted ranked list. We repeat the whole process for all the queries in a separate held-out testing set and report the mean reciprocal rank (MRR)~\cite{voorhees1999trec}, which is the most popular evaluation metric used in measuring the performance of query auto completion tasks.\\
\end{enumerate}

\noindent{\bf Baseline Methods:}
For all the quantitative evaluation tasks, we compare the proposed Joint Influence Model against the obvious baseline method, i.e., \emph{Independent Influence Model} (We call it ``IIM") introduced in~\cite{karmaker2017modeling}. If the Joint Influence Model( \emph{JIM}) performs better than \emph{IIM}, we can conclude that capturing inter-dependencies is indeed useful and can help us achieve superior performance in real life applications.  Additionally, as all these quantitative evaluation tasks are some kind of forecasting problems, we also use some popular time series prediction methods as the baselines including Autoregressive Models (AR), Vector Auto Regression (VAR) etc. Note that, our primary focus is not the quantitative evaluation, rather demonstrating the usefulness of capturing influence inter-dependencies among different events. Thus, experimenting with many different forecasting methods is an orthogonal direction with respect to our focus which we do not explore in this work. We also include the simplest baseline method \emph{Naive Frequency} (NF), where the current hour's frequency is used to predict the next hour's frequency. Table~\ref{table:baselines} lists down all the methods we experimented and also provides with an acronym for each method for notational convenience. \emph{JIM} is the ``Joint Influence Model" proposed in this paper, whereas, ``JIM-G" is a minor variation of ``JIM" with the constraint that events share the same $\alpha$, i.e., the decay parameter.


\begin{table}
	\begin{center}
		\begin{tabular}{|c|l|}\hline
			{\bf Acronym} & \bf{Method}\\\hline
			NF & Naive Frequency\\
			AR & Auto Regression~\cite{chatfield2016analysis}\\
			ARD & Auto Regression with difference~\cite{chatfield2016analysis}\\
			VAR & Vector Auto Regression~\cite{box2015time}\\
			IIM & Independent Influence Model~\cite{karmaker2017modeling}\\
			JIM & Joint Influence Model\\
			JIM-G & Joint Influence Model-Generalized\\\hline
		\end{tabular}
	\end{center}
	\caption{Methods Compared for Quantitative Evaluation}
	\label{table:baselines}
	\vspace{-5mm}
\end{table}

\section{Results}\label{sec:results}

\subsection{Qualitative Evaluation of the Model} \label{sec:resultsquality}
First, we do a qualitative investigation of the optimal parameters learnt through the optimization process as described in section~\ref{sec:estimation}. Table~\ref{table:parameters} presents these learnt parameters. While the individual numbers in Table~\ref{table:parameters} are not very meaningful, the comparison across different categories of events is quite interesting. For example, $\eta$ for ``Sports" category ($0.3170$) is generally much higher than that for ``World" category ($0.0740$), suggesting that the general interest in ``Sports" events is much higher than ``World" events among the common mass. Another interesting parameter is $\alpha$, which indicates the longevity of influence for different categories of events. According to Table~\ref{table:parameters}, ``World" events ($\alpha=0.6770$) usually have a longer lasting influence compared to ``Sports" events ($\alpha=1.1999$). Next, we move onto providing answers to the specific research questions asked in section~\ref{sec:questionsquality}, sequentially one at a time.\\

\begin{table}
    \begin{center}
        \begin{tabular}{ |c|c|c|c|c|c|c|c|c|c|c| }\hline
        parameter & $\eta$ & $\alpha$ & $\rho$ & $\mu$ & $\phi$ & $\psi$\\\hline
	{\bf Movies} & 0.1961 & 0.8697 & 4.9706 & 3.0197 & 0.4542 & 0.1644\\\hline
	{\bf Sports} & 0.317 & 1.1999 & 6.2745 & 4.2272 & 1.1608 & 0.5304\\\hline
	{\bf US} & 0.2328 & 1.0999 & 6.3056 & 1.777 & 0.6962 & 0.508\\\hline
	{\bf World} & 0.074 & 0.677 & 3.9747 & 1.5226 & 0.2465 & 0.1685\\\hline
          \end{tabular}
    \end{center}
    \caption{Parameters learnt for different categories of events}
    \label{table:parameters}
    \vspace{-4mm}
\end{table}

\noindent{\bf Is the ``Query Generation Process'' well defined?}\\
Table~\ref{sec:SpectralRadius} shows the spectral radius of Mutual-Influence Co-efficient Matrix obtained for different categories of events. It is evident that, all the numbers are less than 1. Thus, we conclude that, the ``Query generation process'' 
is indeed well defined.\\

\begin{table}\small
    \begin{center}
        \begin{tabular}{ |c|c|c|c| }\hline
        {\bf Movies} (0.9319)& {\bf Sports} (0.9649)& {\bf US} (0.9192)& {\bf World} (0.9213)\\\hline
        \end{tabular}
    \end{center}
    \caption{Spectral Radius of MIC Mat. for different categories}
    \label{sec:SpectralRadius}
    \vspace{-4mm}
\end{table}

\noindent{\bf How to compare influences posed by different events?}\\
Table~\ref{table:topevents} reports the top 2 influential events from each category along with their average influence score computed by the formula presented in section~\ref{sec:questionsquality}. For example, the movie ``Captain America: Civil War" was found to be the most influential event in the ``Movies" Category with an average influence score of $11.5514$, while ``Donald trump Vs Hillary Clinton" was found to be the most  influential event (average score 14.0117) in the ``US" category. Manual inspection reveals that all these reported influential events are indeed popular events which match with our intuition.

\begin{table*}\small
    \begin{center}
        \begin{tabular}{ |c||p{3.7cm}|p{3.4cm}|p{3.7cm}|p{4cm}| }\hline
         & \multicolumn{4}{c|}{\bf Sections} \\\cline{2-5}
         {\bf Events} & {\bf Movies} & {\bf Sports} & {\bf US} & {\bf World}\\\hline\hline
         $1$ & Movie: ``Captain America: Civil War'' (11.5514) & Horse-Racing: Kentucky Derby (13.5346)  & Donald trump Vs Hillary Clinton (14.0117) & Panama Papers Released (0.8179)\\\hline 
         $2$ & Movie: ``X-men: Apocalypse'' (2.0532) & Basketball: Stephen Curry (6.6432) & Las Vegas Squatters Housing Collapse (9.6340) &  Philippine Presidential Race ( 0.5821)\\\hline 
        \end{tabular}
    \end{center}
    \caption{Top two most influential events from four different Categories}
    \label{table:topevents}
    \vspace{-7mm}
\end{table*}

\noindent{\bf How to compare \emph{Direct} Vs  \emph{Indirect} influence ?}\\
Table~\ref{table:directVsIndirect} reports the average of the diagonal elements (\emph{Direct Influence}) as well as the non-diagonal elements (\emph{Indirect Influence}) of the MIC matrix for each category of events. It is evident that the influence posed by the triggering event, i.e., \emph{Direct Influence} is significantly larger than that of a non-triggering event, i.e., \emph{Indirect Influence} which also concur with our expectation. For example,  \emph{Direct Influence} ($0.6342$) of events in the ``US" category is much higher than the \emph{Indirect Influence} ($0.0201$) in the same category. In fact, this observation holds for any category.\\

\begin{table}
    \begin{center}
        \begin{tabular}{ |c|c|c|c|c| }\hline
        {\bf Influence} & {\bf Movies} & {\bf Sports} & {\bf US} & {\bf World} \\\hline
        Direct & 0.5285 & 0.6495 & 0.6342 & 0.5798\\
        Indirect & 0.0255 & 0.0165 & 0.0201 & 0.0166\\\hline
        \end{tabular}
    \end{center}
    \caption{\emph{Direct Influence} Vs \emph{Indirect Influence}}
    \label{table:directVsIndirect}
    \vspace{-3mm}
\end{table}


\noindent{\bf How to measure the influence longevity of an event?}\\
Direct inspection of $\alpha$ values from Table~\ref{table:parameters} can provide answer to this question. For example, Table~\ref{table:parameters} suggests that ``Sports" events generally have short term influence ($\alpha=1.1999$ ), while ``World" events have comparatively long lasting influence ($\alpha=0.6770$ ).\\

\begin{figure}[!htb]
	\vspace{-3mm}
	\centering
	\includegraphics[width=75.0mm, height=45.0mm]{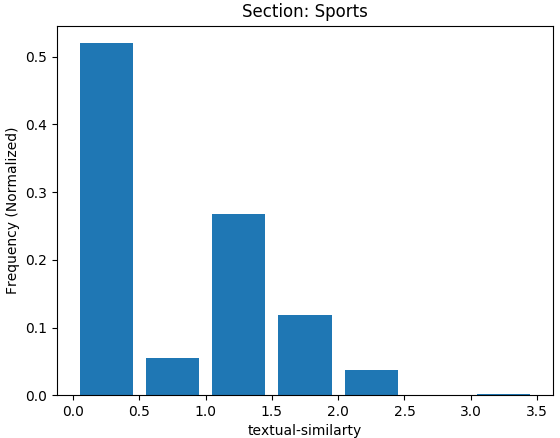}
	\vspace{-4mm}
	\caption{Intent Match Distribution for category ``Sports''}
	\label{fig:pareto}
	\vspace{-2mm}
\end{figure}

\noindent{\bf Is ``Pareto Dist." suitable for ``Intent Match Dist."?}\\
To answer this question, we show the plot for the empirical distribution of ``Intent Match" score between event-text and query-text for the events of ``Sports" category in Figure~\ref{fig:pareto}. This figure demonstrates that as the ``Intent Match" score goes high, the number of queries with corresponding score becomes exponentially smaller, suggesting that, indeed ``Pareto Distribution" is a reasonable candidate for the ``Intent Match Distribution".

\begin{figure}[!htb]
	\vspace{-2mm}
	\centering
	\includegraphics[width=80.0mm,height=50.0mm]{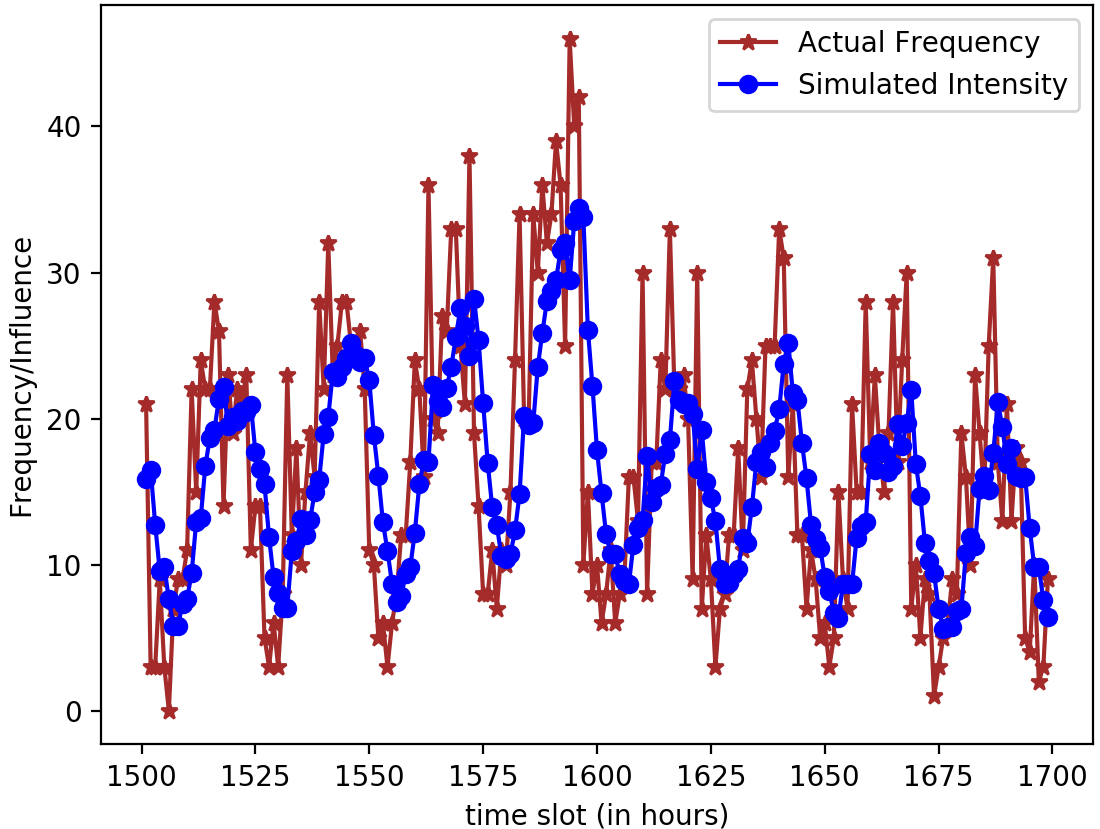}
	\vspace{-2mm}
	\caption{Demonstration of the goodness of fit for the event ``release of movie Captain America: Civil War"}
	\label{fig:goodness}
	\vspace{-2mm}
\end{figure}

\noindent{\bf How well the Model fits the original data?}\\
We plot the the simulated influence of the event ``release of movie Captain America: Civil War" from the ``Movies" category along with the actual frequency of queries generated by that event during the same span of time (hour 1500 to hour 1700) in Figure~\ref{fig:goodness}. It is clearly evident that the simulated influence can indeed capture the trend of the actual frequency of generated queries and thus, we conclude that the model can indeed capture the influence trend with a decent accuracy.

\vspace{-2mm}
\subsection{Applications and Quantitative Evaluation}
This section presents the quantitative evaluation results for the five different application tasks presented in section~\ref{sec:applicaitons}, namely, Predict the most influential event in the future [Table~\ref{table:PredictBestEvent}], Rank multiple events based on their future influences [Table~\ref{table:PredictRankEvent}],  Predict the most frequent query in the future [Table~\ref{table:PredictBestQuery}], Rank queries based on their future frequencies [Table~\ref{table:PredictRankQuery}], Solve a real world application problem, e.g., query auto completion task [Table~\ref{table:PredictAutoComplete}]. General inspection of Table[\ref{table:PredictBestEvent}-\ref{table:PredictAutoComplete}] reveals that, ``JIM-G" is found to be the most robust method for all these different application tasks by obtaining the highest number for performance metrics most of the time. For example, for the task ``Predict the most influential event in the future" [Table~\ref{table:PredictBestEvent}], ``JIM-G" is found to achieve the highest accuracy for all four categories of events. For the ``Query auto completion task", the mean reciprocal rank for ``JIM-G" is found to be the highest for all categories except the category ``World", for which ``IIM" obtains a slightly better number. 

In case of event level predictions (Table~\ref{table:PredictBestEvent} and ~\ref{table:PredictRankEvent}), $JIM$ turns out to be the second best performing method. This suggests that the Joint Influence Model indeed captures useful information which results in its superior performance over other baseline methods. The superiority of ``JIM-G" over ``JIM" may be explained by the fact that, while ``JIM" has more parameters for $\alpha$ (i.e., one $\alpha$ for each single event) than ``JIM-G" (i.e., single $\alpha$ for all events), ``JIM" might be suffering from over-fitting the training data while ``JIM-G" would learn a more general model suitable across multiple events. This over-fitting problem seems more prominent for query level predictions (Table~\ref{table:PredictBestQuery} and ~\ref{table:PredictRankQuery}), especially for category ``World" where the number of queries in the dataset is comparatively very small (Table~\ref{table:JointDataset}). Here, ``JIM" cannot even achieve the second best performance. We believe this is due to the sparsity of query level data. Interestingly, the simple baseline ``NF", achieves quite good result at the query level prediction problems, while ``VAR" suffers severely from overfitting. However, ``JIM-G" still performs the best for most of the cases in query level predictions.

In summary, Table[\ref{table:PredictBestEvent}-\ref{table:PredictAutoComplete}]  suggest that the Joint Influence Model is quite robust and useful in many different applications with superior performance over a number of reasonable baseline methods.

\begin{table}\small
    \begin{center}
       \begin{tabular}{ |c|c|l|l|l|l| }\hline
        {\bf Metric} & {\bf Methods} & {\bf Movies} & {\bf Sports} & {\bf US} & {\bf World} \\\hline
        	  & NF & 0.6638 & 0.6647 &  0.9302 &  0.5073\\
	  & AR & 0.7256 &  0.6818 &  0.8959 &  0.3934\\
	  & ARD & 0.7445 & 0.4249 & 0.9388 & 0.0609\\
	 Accuracy &VAR & 0.7399 & 0.4997 & 0.5105 &  0.1237\\
	  & IIM & 0.7193 & 0.6162 & 0.9376 &  0.5671\textsuperscript{2}\\
	  & JIM & 0.7520\textsuperscript{2} & 0.6938\textsuperscript{2} &  0.9491\textsuperscript{2} &  0.5307\\
	  & JIM-G & {\bf 0.7531\textsuperscript{1}} & {\bf 0.6967\textsuperscript{1}} &  {\bf 0.9542\textsuperscript{1}}  & {\bf 0.5905\textsuperscript{1}} \\\hline
	 \end{tabular}
    \end{center}
    \caption{ Predicting the most influential event in future}
    \label{table:PredictBestEvent}
    \vspace{-2mm}

    \begin{center}
        \begin{tabular}{ |c|c|c|c|c|c| }\hline
        {\bf Metric} &{\bf Methods} & {\bf Movies} & {\bf Sports} & {\bf US} & {\bf World} \\\hline
        & NF & 0.9074 &  0.9105 & 0.9792 & 0.7798 \\
        & AR & 0.9370 & 0.9168 & 0.9460 & 0.6951\\
	& ARD & 0.8604 & 0.7458 & 0.9529 & 0.4358\\
	NDCG& VAR & 0.8831 & 0.7914 & 0.8950 &  0.5175\\
	& IIM & 0.9348 & 0.8975 &  0.9831 &  0.8393\textsuperscript{2}\\
        & JIM & 0.9485\textsuperscript{2} & 0.9278\textsuperscript{2}  & 0.9870\textsuperscript{2} & 0.8275 \\
        & JIM-G & {\bf 0.9508\textsuperscript{1}} & {\bf 0.9322\textsuperscript{1}} &  {\bf 0.9879\textsuperscript{1}} &  {\bf 0.8517\textsuperscript{1}} \\\hline\hline
         & NF & 0.6596 & 0.6800 & 0.8573 & 0.5140\\
         & AR & 0.7052 &  0.6821 & 0.7695 & 0.3967\\
	 & ARD & 0.5320 & 0.4122 & 0.7267 & 0.0942\\
	 RBO & VAR & 0.5752 & 0.4808 & 0.6331 &  0.1647\\
	 & IIM & 0.6961 & 0.6479 &  0.8597 &  0.5992\textsuperscript{2}\\
         & JIM & 0.7194\textsuperscript{2} & 0.6980\textsuperscript{2} &  0.8672\textsuperscript{2} & 0.5623\\
         & JIM-G & {\bf 0.7252\textsuperscript{1}} & {\bf 0.7069\textsuperscript{1}} &  {\bf 0.8705\textsuperscript{1}} & {\bf 0.6087\textsuperscript{1}}\\\hline
        \end{tabular}
    \end{center}
    \caption{ Predicting future influences of multiple events (Wilcoxon's signed rank test at level 0.05)}
    \label{table:PredictRankEvent}
    \vspace{-6mm}
\end{table}

\begin{table}\small
    \begin{center}
       \begin{tabular}{ |c|c|l|l|l|l| }\hline
        {\bf Metric} & {\bf Methods} & {\bf Movies} & {\bf Sports} & {\bf US} & {\bf World} \\\hline
         & NF & 0.3281 & 0.4894\textsuperscript{2} & 0.5717\textsuperscript{2} &  0.3879 \\
         & AR & {\bf 0.3879\textsuperscript{1}} &  0.4794 & 0.5400 &  0.4504\\
	 & ARD & 0.2424 & 0.1965 &  0.4410 & 0.0443\\
	 Accuracy &VAR & 0.0023 & 0.0007 & 0.0029 &  0.0001\\
	 & IIM & 0.3413 & 0.3660 & 0.5408 &  {\bf 0.4710\textsuperscript{1}}\\
         & JIM & 0.3642 & 0.4688 &  0.5563 &  0.3035 \\
        	 & JIM-G & 0.3820\textsuperscript{2} & {\bf 0.5134\textsuperscript{1}} &  {\bf 0.5843\textsuperscript{1}} &  0.4544\textsuperscript{2} \\\hline
	 \end{tabular}
    \end{center}
    \caption{ Predicting the most frequent query in future}
    \label{table:PredictBestQuery}
\vspace{-2mm}
    \begin{center}
        \begin{tabular}{|c|c|c|c|c|c| }\hline
         {\bf Metric} & {\bf Method} & {\bf Movies} & {\bf Sports} & {\bf US} & {\bf World} \\\hline
         & NF & 0.5914 & 0.6693 &  0.8060 &  0.4465 \\
         & AR & 0.6713\textsuperscript{2} & 0.7440\textsuperscript{2}  & 0.7789 & 0.5200\\
	 & ARD & 0.2642 & 0.2977 & 0.4717 & 0.0827\\
	 NDCG & VAR & 0.0087 & 0.0052 &  0.0136 &  0.0015\\
	 & IIM & 0.6355 & 0.6976 & 0.8121\textsuperscript{2}  &  {\bf 0.6555\textsuperscript{1}}\\
         & JIM & 0.6484 & 0.7204 &  0.8022 &  0.4809 \\
         & JIM-G & {\bf 0.6870\textsuperscript{1}} & {\bf 0.7650\textsuperscript{1}} &  {\bf 0.8430\textsuperscript{1}} &  0.6062\textsuperscript{2}  \\\hline\hline
         & NF & 0.4349 & 0.5707 &  0.6491 &  0.3665 \\
         & AR & 0.4947\textsuperscript{2}  & 0.5908\textsuperscript{2}  &  0.6102 & 0.4130\\
	 & ARD & 0.1803 & 0.2191 & 0.3237 & 0.0538\\
	 RBO & VAR & 0.0042 & 0.0019 & 0.0045 &  0.0001\\
	 & IIM & 0.4562 & 0.5174 & 0.6509\textsuperscript{2}  &  {\bf 0.4676\textsuperscript{1}}\\
         & JIM & 0.4782 & 0.5724 &  0.6436 & 0.3048 \\
         & JIM-G & {\bf 0.5059\textsuperscript{1}} & {\bf 0.6172\textsuperscript{1}} &  {\bf 0.6764\textsuperscript{1}} &   0.4332\textsuperscript{2} \\\hline
        \end{tabular}
    \end{center}
    \caption{ Predicting future frequencies for multiple queries. (Wilcoxon's signed rank test at level 0.05)}
    \label{table:PredictRankQuery}
	\vspace{-2mm}
    \begin{center}
       \begin{tabular}{ |c|c|l|l|l|l| }\hline
        {\bf Metric} & {\bf Methods} & {\bf Movies} & {\bf Sports} & {\bf US} & {\bf World} \\\hline
         & NF & 0.6427 & 0.8427 & 0.8489 &  0.6899\\
         & AR & 0.7382 &  0.9129 &  0.8339 & 0.7471\\
	 & ARD & 0.2842 &  0.4077 &  0.5238 &  0.2754\\
	 MRR & VAR & 0.1911 & 0.1722 & 0.1186 &  0.3696\\
	 & IIM & 0.7839 & 0.9171 &  0.8896 &  {\bf 0.9262\textsuperscript{1}}\\
         & JIM & 0.8082\textsuperscript{2} & 0.9509\textsuperscript{2} & 0.8903\textsuperscript{2} & 0.8999\\
         & JIM-G & {\bf 0.8226\textsuperscript{1}} & {\bf 0.9556\textsuperscript{1}} & {\bf 0.8988\textsuperscript{1}} & 0.9193\textsuperscript{2}\\\hline
	 \end{tabular}
    \end{center}
    \caption{Query Auto-Completion Results: MRR reported. (Wilcoxon's signed rank test at level 0.05)}
    \label{table:PredictAutoComplete}
    \vspace{-7mm}
\end{table}

\section{Conclusion}

The assumption that each popular event poses influence upon user search behavior independently is unrealistic as many real world events are closely related to each other. The primary contribution of this paper is to relax this unrealistic assumption made in the previous work by proposing a Joint Influence Model based on multivariate Hawkes Process that captures the inter-dependency of multiple events in terms of the influence posed by them upon user search behavior. Experimental results demonstrate that the proposed method not only effectively capture the temporal dynamics of joint influences by multiple events, but also when applied to various application tasks, achieves superior performance most of the time over different baseline methods that do not consider this mutual-influence among multiple events. This signifies that the mutual influence which exists among multiple correlated events is an important factor which should be considered while designing such influence models.


\section{Acknowledgement}
This material is based upon work supported in part by the National Science Foundation under Grant Numbers
CNS-1408944 , CNS-1513939, and IIS-1629161.

\vspace{-2mm}
\bibliographystyle{ACM-Reference-Format}
\bibliography{sigproc,sigproc2,sigproc4}


\begin{thebibliography}{36}


\ifx \showCODEN    \undefined \def \showCODEN     #1{\unskip}     \fi
\ifx \showDOI      \undefined \def \showDOI       #1{#1}\fi
\ifx \showISBNx    \undefined \def \showISBNx     #1{\unskip}     \fi
\ifx \showISBNxiii \undefined \def \showISBNxiii  #1{\unskip}     \fi
\ifx \showISSN     \undefined \def \showISSN      #1{\unskip}     \fi
\ifx \showLCCN     \undefined \def \showLCCN      #1{\unskip}     \fi
\ifx \shownote     \undefined \def \shownote      #1{#1}          \fi
\ifx \showarticletitle \undefined \def \showarticletitle #1{#1}   \fi
\ifx \showURL      \undefined \def \showURL       {\relax}        \fi
\providecommand\bibfield[2]{#2}
\providecommand\bibinfo[2]{#2}
\providecommand\natexlab[1]{#1}
\providecommand\showeprint[2][]{arXiv:#2}

\bibitem[\protect\citeauthoryear{??}{NYT}{2016}]%
        {NYTimes}
 \bibinfo{year}{2016}\natexlab{}.
\newblock \bibinfo{title}{The Most Popular API: NYtimes Developers Network}.
\newblock
  \bibinfo{howpublished}{\url{https://developer.nytimes.com/most_popular_api_v2.json\#/README}}.
\newblock
\newblock
\shownote{Accessed: 2016-07-30.}


\bibitem[\protect\citeauthoryear{Abdelhaq, Sengstock, and Gertz}{Abdelhaq
  et~al\mbox{.}}{2013}]%
        {abdelhaq2013eventweet}
\bibfield{author}{\bibinfo{person}{Hamed Abdelhaq}, \bibinfo{person}{Christian
  Sengstock}, {and} \bibinfo{person}{Michael Gertz}.}
  \bibinfo{year}{2013}\natexlab{}.
\newblock \showarticletitle{Eventweet: Online localized event detection from
  twitter}.
\newblock \bibinfo{journal}{\emph{Proceedings of the VLDB Endowment}}
  \bibinfo{volume}{6}, \bibinfo{number}{12} (\bibinfo{year}{2013}),
  \bibinfo{pages}{1326--1329}.
\newblock


\bibitem[\protect\citeauthoryear{Atefeh and Khreich}{Atefeh and
  Khreich}{2015}]%
        {atefeh2015survey}
\bibfield{author}{\bibinfo{person}{Farzindar Atefeh} {and}
  \bibinfo{person}{Wael Khreich}.} \bibinfo{year}{2015}\natexlab{}.
\newblock \showarticletitle{A survey of techniques for event detection in
  twitter}.
\newblock \bibinfo{journal}{\emph{Computational Intelligence}}
  \bibinfo{volume}{31}, \bibinfo{number}{1} (\bibinfo{year}{2015}),
  \bibinfo{pages}{132--164}.
\newblock


\bibitem[\protect\citeauthoryear{Berberich, Bedathur, Alonso, and
  Weikum}{Berberich et~al\mbox{.}}{2010}]%
        {berberich2010language}
\bibfield{author}{\bibinfo{person}{Klaus Berberich}, \bibinfo{person}{Srikanta
  Bedathur}, \bibinfo{person}{Omar Alonso}, {and} \bibinfo{person}{Gerhard
  Weikum}.} \bibinfo{year}{2010}\natexlab{}.
\newblock \showarticletitle{A language modeling approach for temporal
  information needs}. In \bibinfo{booktitle}{\emph{European Conference on
  Information Retrieval}}. Springer, \bibinfo{pages}{13--25}.
\newblock


\bibitem[\protect\citeauthoryear{Blundell, Heller, and Beck}{Blundell
  et~al\mbox{.}}{2012}]%
        {Blundell:2012}
\bibfield{author}{\bibinfo{person}{C. Blundell}, \bibinfo{person}{K.~A Heller},
  {and} \bibinfo{person}{J.~M Beck}.} \bibinfo{year}{2012}\natexlab{}.
\newblock \showarticletitle{Modelling Reciprocating Relationships with Hawkes
  Processes}.
\newblock \bibinfo{journal}{\emph{NIPS}} (\bibinfo{year}{2012}).
\newblock


\bibitem[\protect\citeauthoryear{Boggs and Tolle}{Boggs and Tolle}{1995}]%
        {boggs1995sequential}
\bibfield{author}{\bibinfo{person}{Paul~T Boggs} {and} \bibinfo{person}{Jon~W
  Tolle}.} \bibinfo{year}{1995}\natexlab{}.
\newblock \showarticletitle{Sequential quadratic programming}.
\newblock \bibinfo{journal}{\emph{Acta numerica}}  \bibinfo{volume}{4}
  (\bibinfo{year}{1995}), \bibinfo{pages}{1--51}.
\newblock


\bibitem[\protect\citeauthoryear{Box, Jenkins, Reinsel, and Ljung}{Box
  et~al\mbox{.}}{2015}]%
        {box2015time}
\bibfield{author}{\bibinfo{person}{George~EP Box}, \bibinfo{person}{Gwilym~M
  Jenkins}, \bibinfo{person}{Gregory~C Reinsel}, {and} \bibinfo{person}{Greta~M
  Ljung}.} \bibinfo{year}{2015}\natexlab{}.
\newblock \bibinfo{booktitle}{\emph{Time series analysis: forecasting and
  control}}.
\newblock \bibinfo{publisher}{John Wiley \& Sons}.
\newblock


\bibitem[\protect\citeauthoryear{Campos, Dias, Jorge, and Jatowt}{Campos
  et~al\mbox{.}}{2015}]%
        {campos2015survey}
\bibfield{author}{\bibinfo{person}{Ricardo Campos}, \bibinfo{person}{Ga{\"e}l
  Dias}, \bibinfo{person}{Al{\'\i}pio~M Jorge}, {and} \bibinfo{person}{Adam
  Jatowt}.} \bibinfo{year}{2015}\natexlab{}.
\newblock \showarticletitle{Survey of temporal information retrieval and
  related applications}.
\newblock \bibinfo{journal}{\emph{ACM Computing Surveys (CSUR)}}
  \bibinfo{volume}{47}, \bibinfo{number}{2} (\bibinfo{year}{2015}),
  \bibinfo{pages}{15}.
\newblock


\bibitem[\protect\citeauthoryear{Chatfield}{Chatfield}{2016}]%
        {chatfield2016analysis}
\bibfield{author}{\bibinfo{person}{Chris Chatfield}.}
  \bibinfo{year}{2016}\natexlab{}.
\newblock \bibinfo{booktitle}{\emph{The analysis of time series: an
  introduction}}.
\newblock \bibinfo{publisher}{CRC press}.
\newblock


\bibitem[\protect\citeauthoryear{Crane and Sornette}{Crane and
  Sornette}{2008}]%
        {Crane:2008}
\bibfield{author}{\bibinfo{person}{R. Crane} {and} \bibinfo{person}{D.
  Sornette}.} \bibinfo{year}{2008}\natexlab{}.
\newblock \showarticletitle{Robust Dynamic Classes Revealed by Measuring the
  Response Function of a Social System}.
\newblock \bibinfo{journal}{\emph{Proceedings of the National Academy of
  Sciences of the United States of America}} \bibinfo{volume}{105},
  \bibinfo{number}{41} (\bibinfo{year}{2008}), \bibinfo{pages}{15649--15653}.
\newblock


\bibitem[\protect\citeauthoryear{Dakka, Gravano, and Ipeirotis}{Dakka
  et~al\mbox{.}}{2012}]%
        {dakka2012answering}
\bibfield{author}{\bibinfo{person}{Wisam Dakka}, \bibinfo{person}{Luis
  Gravano}, {and} \bibinfo{person}{Panagiotis Ipeirotis}.}
  \bibinfo{year}{2012}\natexlab{}.
\newblock \showarticletitle{Answering general time-sensitive queries}.
\newblock \bibinfo{journal}{\emph{IEEE Transactions on Knowledge and Data
  Engineering}} \bibinfo{volume}{24}, \bibinfo{number}{2}
  (\bibinfo{year}{2012}), \bibinfo{pages}{220--235}.
\newblock


\bibitem[\protect\citeauthoryear{Dong, Mavroeidis, Calabrese, and
  Frossard}{Dong et~al\mbox{.}}{2015}]%
        {dong2015multiscale}
\bibfield{author}{\bibinfo{person}{Xiaowen Dong}, \bibinfo{person}{Dimitrios
  Mavroeidis}, \bibinfo{person}{Francesco Calabrese}, {and}
  \bibinfo{person}{Pascal Frossard}.} \bibinfo{year}{2015}\natexlab{}.
\newblock \showarticletitle{Multiscale event detection in social media}.
\newblock \bibinfo{journal}{\emph{Data Mining and Knowledge Discovery}}
  \bibinfo{volume}{29}, \bibinfo{number}{5} (\bibinfo{year}{2015}),
  \bibinfo{pages}{1374--1405}.
\newblock


\bibitem[\protect\citeauthoryear{Errais, Giesecke, and Goldberg}{Errais
  et~al\mbox{.}}{2010}]%
        {Errais:2010}
\bibfield{author}{\bibinfo{person}{E. Errais}, \bibinfo{person}{K. Giesecke},
  {and} \bibinfo{person}{L.~R. Goldberg}.} \bibinfo{year}{2010}\natexlab{}.
\newblock \showarticletitle{Affine Point Processes and Portfolio Credit Risk}.
\newblock \bibinfo{journal}{\emph{SIAM J. Fin. Math.}} \bibinfo{volume}{1},
  \bibinfo{number}{1} (\bibinfo{date}{Sep} \bibinfo{year}{2010}),
  \bibinfo{pages}{642--665}.
\newblock


\bibitem[\protect\citeauthoryear{Ghoreishi and Sun}{Ghoreishi and Sun}{2013}]%
        {ghoreishi2013predicting}
\bibfield{author}{\bibinfo{person}{Seyyedeh~Newsha Ghoreishi} {and}
  \bibinfo{person}{Aixin Sun}.} \bibinfo{year}{2013}\natexlab{}.
\newblock \showarticletitle{Predicting event-relatedness of popular queries}.
  In \bibinfo{booktitle}{\emph{Proceedings of the 22nd ACM international
  conference on Information \& Knowledge Management}}. ACM,
  \bibinfo{pages}{1193--1196}.
\newblock


\bibitem[\protect\citeauthoryear{Glaudell, Garcia, and Garcia}{Glaudell
  et~al\mbox{.}}{1965}]%
        {glaudell1965nelder}
\bibfield{author}{\bibinfo{person}{Rebecca Glaudell},
  \bibinfo{person}{Rogelio~Tomas Garcia}, {and}
  \bibinfo{person}{Javier~Barranco Garcia}.} \bibinfo{year}{1965}\natexlab{}.
\newblock \showarticletitle{Nelder-Mead Simplex Method}.
\newblock \bibinfo{journal}{\emph{Computer Journal}}  \bibinfo{volume}{7}
  (\bibinfo{year}{1965}), \bibinfo{pages}{308--313}.
\newblock


\bibitem[\protect\citeauthoryear{Hawkes}{Hawkes}{1971}]%
        {hawkes1971spectra}
\bibfield{author}{\bibinfo{person}{Alan~G Hawkes}.}
  \bibinfo{year}{1971}\natexlab{}.
\newblock \showarticletitle{Spectra of some self-exciting and mutually exciting
  point processes}.
\newblock \bibinfo{journal}{\emph{Biometrika}} (\bibinfo{year}{1971}),
  \bibinfo{pages}{83--90}.
\newblock


\bibitem[\protect\citeauthoryear{Jiang, He, and Allan}{Jiang
  et~al\mbox{.}}{2014}]%
        {jiang2014searching}
\bibfield{author}{\bibinfo{person}{Jiepu Jiang}, \bibinfo{person}{Daqing He},
  {and} \bibinfo{person}{James Allan}.} \bibinfo{year}{2014}\natexlab{}.
\newblock \showarticletitle{Searching, browsing, and clicking in a search
  session: changes in user behavior by task and over time}. In
  \bibinfo{booktitle}{\emph{ACM SIGIR}}.
\newblock


\bibitem[\protect\citeauthoryear{Kairam, Morris, Teevan, Liebling, and
  Dumais}{Kairam et~al\mbox{.}}{2013}]%
        {Kairam:2013}
\bibfield{author}{\bibinfo{person}{S. Kairam}, \bibinfo{person}{M. Morris},
  \bibinfo{person}{J. Teevan}, \bibinfo{person}{D. Liebling}, {and}
  \bibinfo{person}{S. Dumais}.} \bibinfo{year}{2013}\natexlab{}.
\newblock \showarticletitle{Towards Supporting Search over Trending Events with
  Social Media}. In \bibinfo{booktitle}{\emph{International AAAI Conference on
  Web and Social Media}}.
\newblock


\bibitem[\protect\citeauthoryear{Kanhabua, Ngoc~Nguyen, and Nejdl}{Kanhabua
  et~al\mbox{.}}{2015}]%
        {kanhabua2015learning}
\bibfield{author}{\bibinfo{person}{Nattiya Kanhabua}, \bibinfo{person}{Tu
  Ngoc~Nguyen}, {and} \bibinfo{person}{Wolfgang Nejdl}.}
  \bibinfo{year}{2015}\natexlab{}.
\newblock \showarticletitle{Learning to detect event-related queries for web
  search}. In \bibinfo{booktitle}{\emph{Proceedings of the 24th International
  Conference on World Wide Web}}. ACM, \bibinfo{pages}{1339--1344}.
\newblock


\bibitem[\protect\citeauthoryear{Karmaker~Santu, Li, Park, Chang, and
  Zhai}{Karmaker~Santu et~al\mbox{.}}{2017}]%
        {karmaker2017modeling}
\bibfield{author}{\bibinfo{person}{Shubhra~Kanti Karmaker~Santu},
  \bibinfo{person}{Liangda Li}, \bibinfo{person}{Dae~Hoon Park},
  \bibinfo{person}{Yi Chang}, {and} \bibinfo{person}{Chengxiang Zhai}.}
  \bibinfo{year}{2017}\natexlab{}.
\newblock \showarticletitle{Modeling the Influence of Popular Trending Events
  on User Search Behavior}. In \bibinfo{booktitle}{\emph{Proceedings of the
  26th International Conference on World Wide Web Companion}}. International
  World Wide Web Conferences Steering Committee, \bibinfo{pages}{535--544}.
\newblock


\bibitem[\protect\citeauthoryear{Kulkarni, Teevan, Svore, and Dumais}{Kulkarni
  et~al\mbox{.}}{2011}]%
        {kulkarni2011understanding}
\bibfield{author}{\bibinfo{person}{Anagha Kulkarni}, \bibinfo{person}{Jaime
  Teevan}, \bibinfo{person}{Krysta~M Svore}, {and} \bibinfo{person}{Susan~T
  Dumais}.} \bibinfo{year}{2011}\natexlab{}.
\newblock \showarticletitle{Understanding temporal query dynamics}. In
  \bibinfo{booktitle}{\emph{Proceedings of the fourth ACM international
  conference on Web search and data mining}}. ACM, \bibinfo{pages}{167--176}.
\newblock


\bibitem[\protect\citeauthoryear{Li, Deng, Dong, Chang, and Zha}{Li
  et~al\mbox{.}}{2014}]%
        {li2014identifying}
\bibfield{author}{\bibinfo{person}{Liangda Li}, \bibinfo{person}{Hongbo Deng},
  \bibinfo{person}{Anlei Dong}, \bibinfo{person}{Yi Chang}, {and}
  \bibinfo{person}{Hongyuan Zha}.} \bibinfo{year}{2014}\natexlab{}.
\newblock \showarticletitle{Identifying and labeling search tasks via
  query-based hawkes processes}. In \bibinfo{booktitle}{\emph{ACM SIGKDD}}.
\newblock


\bibitem[\protect\citeauthoryear{Liniger}{Liniger}{2009}]%
        {liniger2009multivariate}
\bibfield{author}{\bibinfo{person}{Thomas~Josef Liniger}.}
  \bibinfo{year}{2009}\natexlab{}.
\newblock \emph{\bibinfo{title}{Multivariate hawkes processes}}.
\newblock \bibinfo{thesistype}{Ph.D. Dissertation}. \bibinfo{school}{SWISS
  FEDERAL INSTITUTE OF TECHNOLOGY ZURICH}.
\newblock


\bibitem[\protect\citeauthoryear{Matsubara, Sakurai, and Faloutsos}{Matsubara
  et~al\mbox{.}}{2015}]%
        {matsubara2015web}
\bibfield{author}{\bibinfo{person}{Yasuko Matsubara}, \bibinfo{person}{Yasushi
  Sakurai}, {and} \bibinfo{person}{Christos Faloutsos}.}
  \bibinfo{year}{2015}\natexlab{}.
\newblock \showarticletitle{The web as a jungle: Non-linear dynamical systems
  for co-evolving online activities}. In \bibinfo{booktitle}{\emph{WWW, 2015}}.
  \bibinfo{pages}{721--731}.
\newblock


\bibitem[\protect\citeauthoryear{Pekhimenko, Lymberopoulos, Riva, Strauss, and
  Burger}{Pekhimenko et~al\mbox{.}}{2015}]%
        {Pekhimenko:2015}
\bibfield{author}{\bibinfo{person}{G. Pekhimenko}, \bibinfo{person}{D.
  Lymberopoulos}, \bibinfo{person}{O. Riva}, \bibinfo{person}{K. Strauss},
  {and} \bibinfo{person}{D. Burger}.} \bibinfo{year}{2015}\natexlab{}.
\newblock \showarticletitle{PocketTrend: Timely Identification and Delivery of
  Trending Search Content to Mobile Users}. In \bibinfo{booktitle}{\emph{WWW,
  2015}}.
\newblock


\bibitem[\protect\citeauthoryear{Stomakhin, Short, and Bertozzi}{Stomakhin
  et~al\mbox{.}}{2011}]%
        {Stomakhin:2011}
\bibfield{author}{\bibinfo{person}{A. Stomakhin}, \bibinfo{person}{M.~B.
  Short}, {and} \bibinfo{person}{A.~L. Bertozzi}.}
  \bibinfo{year}{2011}\natexlab{}.
\newblock \showarticletitle{Reconstruction of missing data in social networks
  based on temporal patterns of interactions}.
\newblock \bibinfo{journal}{\emph{Inverse Problems.}} \bibinfo{volume}{27},
  \bibinfo{number}{11} (\bibinfo{date}{Nov} \bibinfo{year}{2011}).
\newblock


\bibitem[\protect\citeauthoryear{Str{\"o}tgen and Gertz}{Str{\"o}tgen and
  Gertz}{2012}]%
        {strotgen2012event}
\bibfield{author}{\bibinfo{person}{Jannik Str{\"o}tgen} {and}
  \bibinfo{person}{Michael Gertz}.} \bibinfo{year}{2012}\natexlab{}.
\newblock \showarticletitle{Event-centric search and exploration in document
  collections}. In \bibinfo{booktitle}{\emph{Proceedings of the 12th
  ACM/IEEE-CS joint conference on Digital Libraries}}. ACM,
  \bibinfo{pages}{223--232}.
\newblock


\bibitem[\protect\citeauthoryear{Voorhees et~al\mbox{.}}{Voorhees
  et~al\mbox{.}}{1999}]%
        {voorhees1999trec}
\bibfield{author}{\bibinfo{person}{Ellen~M Voorhees} {et~al\mbox{.}}}
  \bibinfo{year}{1999}\natexlab{}.
\newblock \showarticletitle{The TREC-8 Question Answering Track Report.}. In
  \bibinfo{booktitle}{\emph{Trec}}, Vol.~\bibinfo{volume}{99}.
  \bibinfo{pages}{77--82}.
\newblock


\bibitem[\protect\citeauthoryear{Walther and Kaisser}{Walther and
  Kaisser}{2013}]%
        {walther2013geo}
\bibfield{author}{\bibinfo{person}{Maximilian Walther} {and}
  \bibinfo{person}{Michael Kaisser}.} \bibinfo{year}{2013}\natexlab{}.
\newblock \showarticletitle{Geo-spatial Event Detection in the Twitter
  Stream.}. In \bibinfo{booktitle}{\emph{ECIR}}. Springer,
  \bibinfo{pages}{356--367}.
\newblock


\bibitem[\protect\citeauthoryear{Wang, Wang, Li, He, Chen, and Liu}{Wang
  et~al\mbox{.}}{2013}]%
        {wang2013theoretical}
\bibfield{author}{\bibinfo{person}{Yining Wang}, \bibinfo{person}{Liwei Wang},
  \bibinfo{person}{Yuanzhi Li}, \bibinfo{person}{Di He}, \bibinfo{person}{Wei
  Chen}, {and} \bibinfo{person}{Tie-Yan Liu}.} \bibinfo{year}{2013}\natexlab{}.
\newblock \showarticletitle{A theoretical analysis of NDCG ranking measures}.
  In \bibinfo{booktitle}{\emph{Proceedings of the 26th Annual Conference on
  Learning Theory (COLT 2013)}}.
\newblock


\bibitem[\protect\citeauthoryear{Webber, Moffat, and Zobel}{Webber
  et~al\mbox{.}}{2010}]%
        {webber2010similarity}
\bibfield{author}{\bibinfo{person}{William Webber}, \bibinfo{person}{Alistair
  Moffat}, {and} \bibinfo{person}{Justin Zobel}.}
  \bibinfo{year}{2010}\natexlab{}.
\newblock \showarticletitle{A similarity measure for indefinite rankings}.
\newblock \bibinfo{journal}{\emph{ACM Transactions on Information Systems
  (TOIS)}} \bibinfo{volume}{28}, \bibinfo{number}{4} (\bibinfo{year}{2010}),
  \bibinfo{pages}{20}.
\newblock


\bibitem[\protect\citeauthoryear{White, Chu, Hassan, He, Song, and Wang}{White
  et~al\mbox{.}}{2013}]%
        {white2013enhancing}
\bibfield{author}{\bibinfo{person}{Ryen~W White}, \bibinfo{person}{Wei Chu},
  \bibinfo{person}{Ahmed Hassan}, \bibinfo{person}{Xiaodong He},
  \bibinfo{person}{Yang Song}, {and} \bibinfo{person}{Hongning Wang}.}
  \bibinfo{year}{2013}\natexlab{}.
\newblock \showarticletitle{Enhancing personalized search by mining and
  modeling task behavior}. In \bibinfo{booktitle}{\emph{WWW, 2013}}.
\newblock


\bibitem[\protect\citeauthoryear{Z.-Mangion, Dewarc, Kadirkamanathand, and
  Sanguinetti}{Z.-Mangion et~al\mbox{.}}{2012}]%
        {Zammit-Mangion:2012}
\bibfield{author}{\bibinfo{person}{A. Z.-Mangion}, \bibinfo{person}{M. Dewarc},
  \bibinfo{person}{V. Kadirkamanathand}, {and} \bibinfo{person}{G.
  Sanguinetti}.} \bibinfo{year}{2012}\natexlab{}.
\newblock \showarticletitle{Point process modelling of the Afghan War Diary}.
\newblock \bibinfo{journal}{\emph{PNAS}} \bibinfo{volume}{109},
  \bibinfo{number}{31} (\bibinfo{date}{July} \bibinfo{year}{2012}),
  \bibinfo{pages}{12414--12419}.
\newblock


\bibitem[\protect\citeauthoryear{Zhang, Konda, Dong, Kolari, Chang, and
  Zheng}{Zhang et~al\mbox{.}}{2010}]%
        {zhang2010learning}
\bibfield{author}{\bibinfo{person}{Ruiqiang Zhang}, \bibinfo{person}{Yuki
  Konda}, \bibinfo{person}{Anlei Dong}, \bibinfo{person}{Pranam Kolari},
  \bibinfo{person}{Yi Chang}, {and} \bibinfo{person}{Zhaohui Zheng}.}
  \bibinfo{year}{2010}\natexlab{}.
\newblock \showarticletitle{Learning recurrent event queries for web search}.
  In \bibinfo{booktitle}{\emph{Proceedings of the EMNLP 2010}}. Association for
  Computational Linguistics, \bibinfo{pages}{1129--1139}.
\newblock


\bibitem[\protect\citeauthoryear{Zhou, Chen, and He}{Zhou
  et~al\mbox{.}}{2015}]%
        {zhou2015unsupervised}
\bibfield{author}{\bibinfo{person}{Deyu Zhou}, \bibinfo{person}{Liangyu Chen},
  {and} \bibinfo{person}{Yulan He}.} \bibinfo{year}{2015}\natexlab{}.
\newblock \showarticletitle{An Unsupervised Framework of Exploring Events on
  Twitter: Filtering, Extraction and Categorization.}. In
  \bibinfo{booktitle}{\emph{AAAI}}. \bibinfo{pages}{2468--2475}.
\newblock


\bibitem[\protect\citeauthoryear{Zhuang, Ogata, and Jones}{Zhuang
  et~al\mbox{.}}{2002}]%
        {Zhuang:2002}
\bibfield{author}{\bibinfo{person}{J. Zhuang}, \bibinfo{person}{Y. Ogata},
  {and} \bibinfo{person}{David~V. Jones}.} \bibinfo{year}{2002}\natexlab{}.
\newblock \showarticletitle{Stochastic Declustering of Space-Time Earthquake
  Occurrences}.
\newblock \bibinfo{journal}{\emph{Journal of the American Statistical
  Association.}} \bibinfo{volume}{97}, \bibinfo{number}{458}
  (\bibinfo{year}{2002}), \bibinfo{pages}{369--380}.
\newblock


\end{thebibliography}

\end{document}